\documentclass[conference]{IEEEtran}
\IEEEoverridecommandlockouts
\usepackage[T1]{fontenc}
\usepackage{cite}
\usepackage{amsmath,amssymb,amsfonts}
\usepackage{booktabs}
\usepackage{graphicx}
\usepackage{textcomp}
\usepackage{xcolor}
\def\BibTeX{{\rm B\kern-.05em{\sc i\kern-.025em b}\kern-.08em
    T\kern-.1667em\lower.7ex\hbox{E}\kern-.125emX}}

\begin{document}

\title{Characterizing Treatment-Context Medication Evidence Across Clinic Notes and Structured EHR Medication History}

\author{
\IEEEauthorblockN{Mingyang Jiang\IEEEauthorrefmark{1},
Congning Ni\IEEEauthorrefmark{2},
Weixin Liu\IEEEauthorrefmark{1}, and
Zhijun Yin\IEEEauthorrefmark{2}}
\IEEEauthorblockA{\IEEEauthorrefmark{1}Vanderbilt University, Nashville, TN, USA}
\IEEEauthorblockA{\IEEEauthorrefmark{2}Department of Biomedical Informatics,
Vanderbilt University Medical Center, Nashville, TN, USA}
\IEEEauthorblockA{\{mingyang.jiang, weixin.liu\}@vanderbilt.edu; \{congning.ni.1, zhijun.yin\}@vumc.org}
}

\maketitle

\begin{abstract}
Clinic notes may describe medication changes that are not reflected in structured electronic health record (EHR) medication history at the same encounter.
Differences in terminology and documentation timing further complicate the interpretation of these discrepancies.
We evaluated medication normalization in clinic notes and examined how semantic and temporal matching affected overlap with structured history.
A note-grounded reference was constructed from large language model (LLM)-proposed mentions, deterministic canonical projection, and existing targeted correction records; random-audit records were summarized separately.
On 5{,}360 reference mention rows from held-out patients, conditional canonical agreement increased from 81.25\% with surface matching to 93.58\% after lexical cleanup and curated aliases.
Candidate--reference mention alignment was evaluated separately, yielding an F1 score of 0.6547.
Across 27{,}552 reference rows, exact same-visit overlap was 16.71\%.
Overlap reached 90.22\% when ingredient and broad-category matches were permitted at the same visit or within 30 days before or after the note.
Incomplete mapping of note-derived labels contributed to residual nonoverlap in ontology-backed comparisons.
Exact-label comparison captured only part of the related medication information across sources; broader matches do not establish equivalent medication use or treatment events.
\end{abstract}


\section{Introduction}
Medication information relevant to treatment decisions appears in both structured electronic health records (EHRs) and narrative clinic notes.
Outpatient notes often record medication starts, stops, holds, regimen changes, and formulation details that may not appear in structured medication history at the same encounter.
An absent exact match may therefore reflect differences in terminology, clinical content, or documentation timing.
Comparing the two sources requires consistent medication labels and explicit matching criteria.

Medication extraction and normalization from clinical narratives are established clinical natural language processing (NLP) tasks.
Systems such as MedEx and MedEx-Unstructured Information Management Architecture (MedEx-UIMA), together with evaluations such as the Informatics for Integrating Biology and the Bedside (i2b2) medication extraction challenge, have addressed the identification and standardization of medication information in narrative text \cite{xu2010medex,jiang2014medexuima,uzuner2010extracting}.
MedXN, medExtractR, and staged extraction-plus-harmonization workflows have extended these approaches for research use \cite{sohn2014medxn,weeks2020medextractr,almeida2021twostage}.
An extracted medication mention without a counterpart in structured history is not necessarily an extraction error.
Prior comparisons of narrative and structured medication information have identified discrepancies associated with workflow, timing, and medication reconciliation practices \cite{turchin2009comparison,wang2015prescription}.
Using structured history as the sole reference can therefore confound note-extraction errors with documentation differences, particularly when a note records a treatment change absent from the structured record at the same encounter.

Normalization can itself produce apparent disagreement when equivalent medication expressions map to different labels.
Brand and generic names, formulation variants, abbreviations, and missing synonyms complicate canonical mapping.
RxNorm provides a standardized nomenclature for clinical drugs \cite{nelson2011rxnorm}, but ambiguity and incomplete vocabulary coverage can limit concept normalization even when a mention is correctly identified \cite{newmangriffis2021ambiguity}.
Ingredient- and therapeutic-category matches can identify related medication information across sources without establishing exact medication identity.
Canonical label agreement and cross-source overlap therefore require separate evaluation.

A note-grounded reference allows normalization to be evaluated against note content rather than structured medication history.
Large language models (LLMs) can propose medication mentions for reference construction.
Recent studies have used LLMs to extract information from unstructured EHRs \cite{ntinopoulos2025llmextract} and treatment discontinuation information from patient-generated text \cite{song2025using}.
Related work has examined auditable clinical concept representations \cite{ni2026auditable} and longitudinal clinical event verification \cite{lin2026graphrag}.
Model-generated mentions can contain errors \cite{zhang2024pitfalls,wang2024applications}; the resulting labels should not be treated as independently validated annotations.

We evaluated note-side normalization and cross-source overlap in a clinic-note cohort.
We constructed a reference from LLM-proposed medication mentions, deterministic canonical projection, and existing targeted correction records, and summarized random-audit records separately.
First, we evaluated conditional canonical agreement on reference rows from held-out patients, comparing lexical cleanup and curated aliases with surface matching, generic lexical retrieval, and public ontology-based comparators.
Second, using the corrected full-cohort reference, we characterized overlap with structured medication history using exact, ingredient, broad-category, and temporal matching criteria, with additional analyses of ontology mapping coverage and residual nonoverlap.
The normalization analysis measures note-side label agreement, while the overlap analysis describes relationships between documentation sources.
Neither establishes agreement on medication use or treatment events.

\section{Materials and Methods}
\begin{figure*}[!t]
\centering
\includegraphics[width=0.85\textwidth]{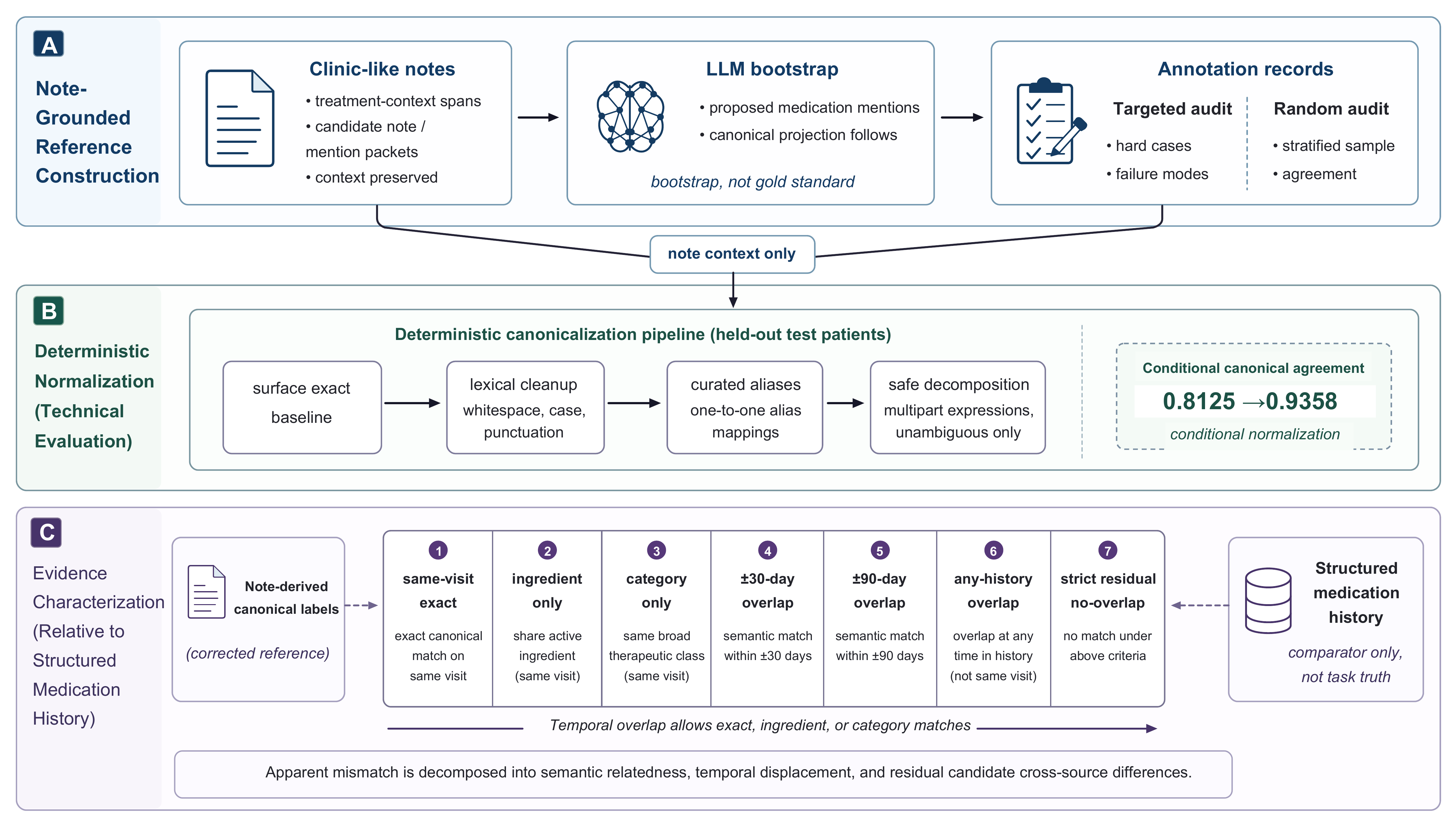}
\vspace{-0.65em}
\caption{Clinic-note packets are converted into an auditable note-grounded reference set through Qwen2.5-32B-Instruct bootstrap output, deterministic projection, and recorded annotation corrections. Deterministic canonicalization is then evaluated on held-out test patients, after which note-derived labels are compared with structured medication history through downstream semantic-temporal characterization.}
\label{fig:workflow}
\end{figure*}

Figure~\ref{fig:workflow} summarizes the workflow.

\subsection{Cohort, note grounding, and audit design}
We conducted a clinic-note-only evaluation of treatment-context medication normalization using a frozen cohort run of 11{,}812 eligible visits from 761 patients and 22{,}483 manifest notes. We restricted the analysis to clinic-like notes because we focused on outpatient documentation of medication starts, stops, holds, regimen changes, and formulation details. The cohort was not filtered on extraction success, preserving a fixed denominator for candidate generation and normalization. This study was determined to be exempt from human subjects research by the Institutional Review Board at Vanderbilt University Medical Center.

The analysis uses several linked units. A \emph{manifest note} was a retained clinic-like note; a \emph{packet note} was a manifest note represented in the adjudication packet layer; a \emph{packet mention} was an upstream candidate mention carried forward for review or reference construction; and a \emph{reference mention row} was a packet-note-based, note-grounded mention row with a non-empty project-specific canonical medication label. The primary evaluation unit was the mention row, and the primary endpoint was mention-level exact canonical agreement with the note-grounded reference label.

Before reference construction, we organized clinic-note snippets into candidate packets for review. We identified candidate spans using medication and treatment-action patterns, seeded medication strings, and cues produced by an upstream medspaCy-based clinical text processing module \cite{eyre2021medspacy}. These candidates were generated from clinic-note context rather than from structured EHR medication fields. In the frozen run, this process produced 10{,}543 packet notes and 31{,}282 packet mentions. The remaining 11{,}940 manifest notes did not enter the packet layer. Their medication content was not exhaustively annotated, so recall over all manifest notes and mentions cannot be estimated from this reference.

Reference mention rows were constructed from clinic-note context through an LLM-bootstrapped, note-grounded workflow. For each packeted clinic-note context, Qwen2.5-32B-Instruct \cite{qwen2025technicalreport}, deployed locally through vLLM \cite{kwon2023pagedattention}, generated schema-constrained JSON output under deterministic decoding (temperature $=0$). For this medication study, we retained only the medication-output branch. The model proposed note-level medication mention strings while excluding dosage, route, and frequency. Medication identity was the normalization target; action attribution was secondary. Agreement on a drug label does not establish agreement on starting, stopping, or changing a regimen. These proposed strings were then normalized, deduplicated within note, aligned to packet mentions, and projected into project-specific canonical medication labels. The resulting resource is an auditable note-grounded reference layer, but it remains packet-note-based rather than a fully independent span-annotated corpus over all manifest notes.

Existing targeted annotation records covered 3{,}388 rows selected for likely errors or ambiguity. The recorded decisions removed 200 rows and retained 3{,}188, comprising 11.57\% of the corrected reference. The remaining 24{,}364 rows retained bootstrap-derived labels after projection into the project canonical label space. A separate stratified random audit sampled 1{,}000 previously unaudited rows by note type, candidate category, and action cue. Its protocol specified bounded note context without structured medication fields. The available annotation records do not establish whether these decisions were made by independent human reviewers.

\subsection{Held-out evaluation and normalization analysis}
All evaluation summaries use the reference after applying the existing 200 row removals and 3{,}148 retained-row canonical-label corrections. Patient assignments, candidate predictions, normalization outputs, and terminology resources were kept fixed.

Reference rows were aligned back to extracted candidate rows to construct a mention-level alignment table. A matched extraction and reference pair was counted as a true positive (TP), a reference row without a matched extracted candidate as a false negative (FN), and an extracted candidate without a matched reference row as a false positive (FP). Structured-EHR comparability flags were retained only for downstream secondary analysis.

We used a patient-level held-out split to limit within-cohort adaptation. The 27{,}552 reference mention rows came from 756 unique patients; these patients were split once into 606 development patients and 150 held-out test patients using a fixed random seed, with approximate balance on patient-level mention-row volume and whether a patient contributed targeted-audit rows. We calculated extraction precision, recall, F1 score, canonicalization performance, comparator results, and alias-refinement sensitivity only on held-out test patients. Reference projection and the evaluated normalizer share project terminology resources. The patient split separates patients but does not establish independence of the base alias map from test-patient material, because its construction chronology is not fully documented. Full-cohort descriptive analyses, including cohort grounding, targeted audit, random audit, and the primary semantic-temporal ladder, were retained as characterization results.

We evaluated the deterministic canonicalization pipeline and its calibration comparators on the held-out test split of 5{,}360 reference mention rows with non-empty project-specific canonical medication labels. The deterministic canonicalization pipeline consisted of surface-exact comparison, lexical cleanup, curated alias mapping, and conservative safe decomposition. The surface-exact baseline compared normalized raw mention strings directly with reference canonical labels; lexical cleanup standardized whitespace, punctuation, and token-level variants; curated alias mapping applied one-to-one deterministic aliases while excluding known ambiguous mappings; and safe decomposition accepted only unambiguous multipart expressions. This endpoint measures mention-level canonicalization among note-grounded reference rows and does not measure end-to-end extraction performance. To calibrate the deterministic gain, we also evaluated public Observational Medical Outcomes Partnership (OMOP)/RxNorm synonym matching without project-specific alias refinement and generic nearest-canonical lexical retrieval based on character $n$-grams, edit similarity, and token overlap. As one recognizable external baseline, we also ran untuned MedXN \cite{sohn2014medxn} on held-out notes and compared note-level recovery of unique note and normalized-medication-label pairs on a predefined treatment-context subset.

\subsection{Semantic-temporal characterization and secondary analyses}
Secondary concordance analyses used exact, ingredient, and broad-category mapping levels. Exact level compared project-specific canonical labels directly. Ingredient level collapsed note-side and structured-side labels to normalized ingredients using deterministic RxNorm-style ingredient mapping where possible, supplemented by project-specific ingredient aliases. Broad-category level grouped normalized medications into project-specific clinical categories for semantic comparison and interpretability.

To make the structured-side comparison more reproducible, we also built explicit OMOP/RxNorm-backed mapping artifacts from local OMOP vocabulary tables, including the core concept, relationship, ancestor, drug-strength, and concept-synonym tables. Broad category assignment relied on ancestor relationships to vocabularies such as the World Health Organization (WHO) Anatomical Therapeutic Chemical (ATC) system \cite{who2026atcddd} and Veterans Affairs (VA) drug classes as represented in National Drug File--Reference Terminology (NDF-RT)-style terminology resources \cite{pathak2011ndfrt}, as available in the OMOP vocabulary dump \cite{reich2024ohdsi}. Structured drug concepts were bridged to standard concept names, ingredient representations, and broad therapeutic categories. In parallel, project canonical labels were mapped to public OMOP/RxNorm concepts with provenance-aware coverage summaries and unmapped-label lists. We used this mapping as a sensitivity analysis alongside the broader note-grounded comparison.

We used a semantic-temporal ladder to compare each note-derived medication row with structured medication history. Same-visit overlap was checked first in descending semantic specificity: exact canonical label, ingredient or generic overlap, and broad therapeutic-category overlap. If no same-visit overlap existed, the same hierarchy was applied within patient-level $\pm 30$-day and $\pm 90$-day windows using note date as the primary anchor and visit start date as fallback, followed by any-history overlap. Temporal buckets can include exact, ingredient, or category matches. Project-level same-visit matching requires a shared visit identifier; the ontology-backed variant additionally requires temporal overlap with the note anchor. Category matches may link different drugs, and broader windows may link separate treatment episodes. We summarized these checks in seven mutually exclusive categories.

\subsection{Evaluation metrics}
The primary normalization endpoint was mention-level exact canonical agreement with the note-grounded reference label, defined as the proportion of reference mention rows whose predicted canonical label exactly matched the note-grounded target. Incremental stage contribution was reported as the change in exact agreement relative to the previous stage. For the secondary visit-level sensitivity analysis, each visit was represented as a set of unique canonical medication labels, and concordance was summarized by exact set match and mean Jaccard overlap.

Residual deterministic-pipeline failures were assigned to interpretable categories, including missing alias, lab/substance or non-medication, combination or formulation mismatch, and ambiguous abbreviation. Frequent unresolved raw mentions were summarized to identify high-yield alias expansion targets. For the historical pre-correction OMOP-backed no-overlap bucket, a sample of recorded annotations was used to distinguish note-side mapping failure, structured-side undercapture, extraction or reference noise, and residual candidate note-only evidence. We also ran a bounded alias-supplement sensitivity in which a transparent supplement derived only from development-patient reviewed note-side mapping failures was added to the deterministic alias map and then evaluated on held-out test patients. Structured-medication-history concordance was treated only as a downstream comparison and never as extraction truth.

\section{Results}
\noindent We report recorded annotation agreement, held-out normalization performance, semantic and temporal comparisons with structured medication history, and analyses of the remaining no-overlap cases.

\subsection{Cohort grounding and reference denominators}
The frozen cohort run included 11{,}812 eligible visits, 22{,}483 manifest notes, 10{,}543 packet notes, 31{,}282 packet mentions, and 27{,}552 reference mention rows with non-empty canonical labels. The targeted audit layer covered 3{,}188 rows (11.57\%). These counts describe the full-cohort reference universe for descriptive analyses; held-out normalization analyses use 5{,}360 test reference mention rows.

\subsection{Reference credibility and targeted audit}
The targeted annotation queue contained 3{,}388 rows, of which 200 were removed and 3{,}188 retained. The retained rows represented 11.57\% of the corrected 27{,}552-row reference. This queue was selected for likely failures and does not estimate reference quality across the cohort.

Among the 3{,}188 retained targeted rows, 40 labels were unchanged and 3{,}148 were corrected. The 1.25\% pre-correction agreement describes the downstream projected labels in this selected queue, rather than Qwen's overall medication extraction accuracy. The existing records attributed 2{,}335 disagreements to candidate-generation miss, 778 to missing alias, 27 to formulation or salt variant, and 8 to combination ingredient mismatch. These are recorded categories rather than independently validated error assignments.

In the random-audit records, 582 of 1{,}000 sampled rows were classified as valid medication mentions with acceptable spans; 418 were not classified as evaluable. Canonical labels agreed on 536/582 evaluable rows (92.10\%), and action labels agreed on 310/582 (53.26\%). Across all 1{,}000 sampled rows, 536 (53.60\%) had recorded canonical agreement. Reporting both denominators distinguishes agreement among evaluable rows from the fraction of the sampled records with canonical agreement.

The targeted records document corrections in a selected failure-enriched subset; the random records summarize agreement conditional on evaluability. Neither establishes an independently annotated full-note reference or independent inter-annotator agreement.

\subsection{Deterministic canonicalization pipeline}
\begin{figure}[!tb]
\centering
\includegraphics[width=\columnwidth]{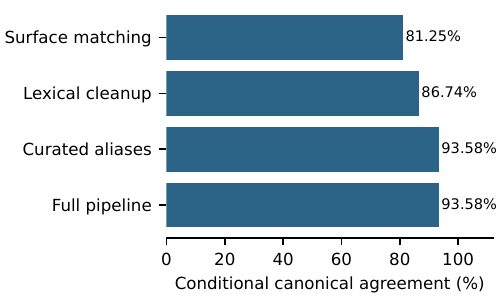}
\caption{Held-out conditional canonical agreement at each deterministic normalization stage.}
\label{fig:norm_ladder}
\end{figure}

Held-out mention alignment yielded 3{,}815 true positives, 2{,}480 false positives, and 1{,}545 false negatives (precision 0.6060, recall 0.7118, F1 0.6547). These scores describe candidate--reference alignment, without requiring a correct canonical label. We then evaluated the conditional canonicalization ladder over 5{,}360 test-split reference mention rows (Table~\ref{tab:norm_ladder} and Fig.~\ref{fig:norm_ladder}). This ladder evaluates surface-exact matching, lexical cleanup, curated alias mapping, and the full deterministic pipeline sequentially on the same fixed held-out denominator. Canonicalization scores are conditional on the constructed reference rows and do not measure end-to-end extraction over all manifest notes. For canonicalization, surface-exact matching achieved exact agreement of 0.8125, corresponding to 4{,}355 correctly normalized mention rows. Lexical cleanup increased agreement to 0.8674, solving 294 additional rows. Curated alias mapping further increased agreement to 0.9358, solving another 367 rows. The full deterministic pipeline remained at 0.9358, indicating that safe decomposition did not provide additional measurable gain on held-out patients.

We compared the pipeline with public-ontology and generic lexical baselines on the same held-out rows (Table~\ref{tab:normalization_comparators}). Generic nearest-canonical lexical comparators remained close to lexical-cleanup performance, with exact canonical agreement from 0.8632 to 0.8634. A public OMOP/RxNorm synonym baseline that avoided project-specific alias refinement was lower still: 0.7416 when preserving the matched public term surface and 0.6493 under stricter ingredient-style collapse. These results suggest that the main gain of the deterministic pipeline is not explained by generic fuzzy matching alone; it depends on clinic-note-specific normalization coverage that remains transparent and auditable.

For external positioning, we summarized held-out end-to-end note-level recovery of unique note and normalized-medication-label pairs on the predefined treatment-context subset (Table~\ref{tab:external_positioning}). This subset contained 1{,}073 held-out notes and 2{,}585 reference note-label pairs linked to action-bearing treatment-context rows. On the full treatment-context subset, the local extraction plus deterministic canonicalization pipeline achieved precision 0.7800, recall 0.7404, and F1 0.7597. Untuned MedXN produced at least one prediction for 514 of these treatment-context notes. On this MedXN-output subset, the local pipeline achieved precision 0.7754, recall 0.7556, and F1 0.7654, whereas MedXN achieved precision 0.3628, recall 0.7904, and F1 0.4973. Untuned MedXN had higher recall on the common subset (0.7904 versus 0.7556), while the local pipeline had higher precision (0.7754 versus 0.3628). Both systems were scored against project-specific canonical labels, and the common subset excludes notes with no MedXN output. These vocabulary and subset choices limit generalization of the comparison beyond the reported setting.

\begin{table}[t]
\caption{Held-out conditional canonicalization ablation. Lexical cleanup and curated aliases explain nearly all measurable gain on held-out patients.}
\label{tab:norm_ladder}
\centering
\footnotesize
\setlength{\tabcolsep}{3pt}
\begin{tabular}{lrrr}
\toprule
\textbf{Stage} & \textbf{Agreement} & \textbf{$\Delta$} & \textbf{Rows (\%)} \\
\midrule
Surface-exact baseline & 0.8125 & 0.0000 & 4{,}355 (81.25\%) \\
Lexical cleanup & 0.8674 & 0.0549 & 294 (5.49\%) \\
Curated alias map & 0.9358 & 0.0685 & 367 (6.85\%) \\
\shortstack[l]{Full deterministic\\pipeline} & 0.9358 & 0.0000 & 0 (0.00\%) \\
\shortstack[l]{Unresolved after\\full pipeline} & --- & --- & 344 (6.42\%) \\
\bottomrule
\end{tabular}
\end{table}

\begin{table*}[t]
\caption{Held-out calibration comparators on the same 5{,}360-row denominator. These methods contextualize the deterministic canonicalization pipeline rather than serving as a definitive benchmark.}
\label{tab:normalization_comparators}
\centering
\footnotesize
\setlength{\tabcolsep}{4pt}
\begin{tabular}{p{0.28\textwidth}p{0.28\textwidth}rr}
\toprule
\textbf{Method} & \textbf{Knowledge source / rule type} & \shortstack{\textbf{Exact canonical}\\\textbf{agreement}} & \shortstack{\textbf{Prediction}\\\textbf{rate}} \\
\midrule
Surface-exact baseline & none & 0.8125 & 1.0000 \\
Lexical cleanup & deterministic cleanup & 0.8674 & 1.0000 \\
OMOP/RxNorm synonym, preserve term & public ontology only & 0.7416 & 0.8653 \\
OMOP/RxNorm synonym, ingredient & public ontology only & 0.6493 & 0.8625 \\
Char $n$-gram retrieval & generic string retrieval & 0.8634 & 1.0000 \\
Edit retrieval & generic string retrieval & 0.8632 & 1.0000 \\
Token-Jaccard retrieval & generic string retrieval & 0.8634 & 1.0000 \\
Full deterministic pipeline & project-specific deterministic aliases & 0.9358 & 1.0000 \\
\bottomrule
\end{tabular}
\vspace{-0.2em}
\end{table*}

\begin{table*}[t]
\caption{Held-out treatment-context external positioning using unique note and normalized-medication-label pairs. The treatment-context subset was defined a priori from held-out rows with action-bearing cues such as start, stop, hold, and dose change. MedXN was run untuned with chunked-note export and is reported on the subset of treatment-context notes that yielded at least one MedXN output; the local pipeline is rescored on the same subset for comparability.}
\label{tab:external_positioning}
\centering
\footnotesize
\setlength{\tabcolsep}{3.5pt}
\begin{tabular}{p{0.30\textwidth}rrrrrrrrr}
\toprule
\textbf{System} & \textbf{Notes} & \shortstack{\textbf{Reference}\\\textbf{pairs}} & \shortstack{\textbf{Predicted}\\\textbf{pairs}} & \textbf{TP} & \textbf{FP} & \textbf{FN} & \textbf{Precision} & \textbf{Recall} & \textbf{F1} \\
\midrule
Local extraction + deterministic canonicalization (treatment-context subset) & 1{,}073 & 2{,}585 & 2{,}454 & 1{,}914 & 540 & 671 & 0.7800 & 0.7404 & 0.7597 \\
Local extraction + deterministic canonicalization (MedXN-output treatment-context subset) & 514 & 1{,}293 & 1{,}260 & 977 & 283 & 316 & 0.7754 & 0.7556 & 0.7654 \\
MedXN untuned external baseline (MedXN-output treatment-context subset) & 514 & 1{,}293 & 2{,}817 & 1{,}022 & 1{,}795 & 271 & 0.3628 & 0.7904 & 0.4973 \\
\bottomrule
\end{tabular}
\end{table*}

\subsection{Semantic-temporal mismatch characterization}
\begin{figure}[!t]
\centering
\includegraphics[width=\columnwidth]{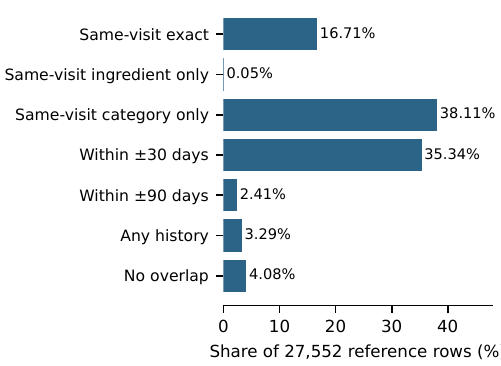}
\caption{Semantic-temporal mismatch ladder. Bars are mutually exclusive and characterize how note-derived medication evidence relates to structured medication history rather than the truth of medication use.}
\label{fig:temporal_mismatch_ladder}
\end{figure}

We next examined how broader semantic and temporal matching criteria changed overlap with structured medication history. Figure~\ref{fig:temporal_mismatch_ladder} summarizes the seven-bucket semantic-temporal ladder. Same-visit exact overlap remained uncommon at 4{,}605/27{,}552 = 16.71\%. Many rows were recovered by semantic or temporal relaxation, including same-visit category-only overlap at 10{,}501/27{,}552 = 38.11\% and $\pm 30$-day overlap after failing same-visit at 9{,}737/27{,}552 = 35.34\%. Only 1{,}125/27{,}552 = 4.08\% remained in the strict no-structured-overlap bucket under broad project-level mapping.

Cumulatively, 54.88\% of note-derived mention rows had same-visit semantic overlap, 90.22\% had same-visit or $\pm 30$-day overlap, and 95.92\% had some structured-history overlap. Most note-derived medication rows had a related structured-history entry when broader semantic matching and the $\pm 30$-day window were permitted.

Low same-visit exact overlap increased under broader semantic and temporal criteria. Much of the increase arose from category-level matches, which indicate related therapeutic content but may involve different drugs. The ladder describes how overlap changes with comparison granularity; it does not establish correspondence between medication exposures or treatment decisions.

The OMOP/RxNorm-backed sensitivity analysis examined how note-side mapping coverage affected strict no-overlap. Structured medication history standardized well under OMOP/RxNorm, with 5{,}149/5{,}172 unique structured drug concepts (99.56\%) bridged to standard concepts, covering 92.52\% of structured medication rows. Note-side public ontology coverage was much lower at the unique-label level: only 657/2{,}124 canonical labels (30.93\%) mapped cleanly to public OMOP/RxNorm concepts. These mapped labels accounted for 18{,}411/27{,}552 note-derived mention rows (66.82\%), showing that mapping coverage was higher for frequently occurring labels.

Ingredient coverage reached 66.63\% of rows and broad-category coverage reached 66.26\%. The 4.08\% broad-ladder no-structured-overlap bucket and the 41.05\% strict OMOP/RxNorm no-overlap sensitivity in Table~\ref{tab:no_overlap_calibration} therefore answer different questions: the first is a project-level semantic-temporal characterization result, whereas the second applies stricter ontology-backed comparison rules. Under these stricter rules, no-overlap was 11.78\% among mapped note rows and 11.26\% among rows with both ingredient and category coverage. By provenance, all 9{,}141 ontology-unmapped note rows remained in the strict no-overlap bucket, whereas only 2{,}169 of 18{,}411 public-ontology-mapped rows (11.78\%) did so. The lower mismatch rate among mapped rows suggests that incomplete note-side ontology coverage contributes substantially to strict nonoverlap.

Recorded annotations for 100 pre-correction OMOP-backed no-overlap cases classified 80 as note-side mapping failure, 7 candidate note-only evidence, 6 structured-side undercapture, and 7 extraction or reference noise. These classifications predate the reference correction and do not estimate the composition of the corrected residual bucket. To assess mapping sensitivity on held-out patients, we rebuilt the supplement only from development-patient reviewed failures and reapplied it to held-out test patients. On 5{,}360 held-out test rows, mapped-row coverage increased from 0.6823 to 0.7498, and the strict OMOP-backed no-overlap bucket fell from 2{,}096 rows (39.10\%) to 1{,}752 rows (32.69\%). The supplement increased mapped coverage by 362 rows and reduced no overlap by 344 rows under the fixed comparison rules. Table~\ref{tab:no_overlap_calibration} summarizes strict no-overlap and mapping sensitivity. Strict no-overlap can reflect mapping gaps, documentation differences, or extraction and reference errors.

\subsection{Semantic note-to-structured-record concordance remains secondary}
Structured medication history was retained only as a secondary concordance analysis. To test whether the two documentation layers were related despite low exact agreement, we summarized visit-level concordance across exact, ingredient, and broad-category semantic levels. Across 5{,}320 note--structured-record visit pairs, exact-set agreement and overlap were low at exact and ingredient levels, with exact-set-match rate 0.0058 at both levels, overlap rate 0.1898 at exact canonical level, overlap rate 0.1908 at ingredient level, and mean Jaccard values near 0.043.

\begin{table}[t]
\caption{Secondary semantic concordance increases at broader clinical-category level, supporting semantic relatedness without equivalence.}
\label{tab:semantic_concordance}
\centering
\footnotesize
\begin{tabular}{lccc}
\toprule
\textbf{Semantic level} & \textbf{Set match} & \textbf{Overlap} & \textbf{Jaccard} \\
\midrule
Exact canonical & 0.0058 & 0.1898 & 0.0430 \\
Ingredient level & 0.0058 & 0.1908 & 0.0433 \\
Broad category & 0.1665 & 0.4940 & 0.3068 \\
\bottomrule
\end{tabular}
\end{table}

At broad clinical-category level, concordance was materially higher: overlap rate 0.4940, exact-set-match rate 0.1665, and mean Jaccard 0.3068. The higher overlap and Jaccard values at the broad-category level indicate that the sources often refer to similar medication classes despite low exact agreement.

\begin{table*}[!t]
\caption{Ontology-backed no-overlap and mapping sensitivity. Rows 1--2 use the corrected full cohort; row 3 describes a pre-correction sample; rows 4--5 use held-out test patients.}
\label{tab:no_overlap_calibration}
\centering
\footnotesize
\begin{tabular}{p{0.07\textwidth}p{0.27\textwidth}p{0.27\textwidth}p{0.23\textwidth}}
\toprule
\textbf{Step} & \textbf{Analysis} & \textbf{Result} & \textbf{Interpretation} \\
\midrule
1 & Strict OMOP/RxNorm no-overlap & 41.05\% & \shortstack[l]{High no-overlap under\\strict mapping} \\
2 & Mapped-note subset & 11.78\% & \shortstack[l]{Lower no-overlap among\\mapped rows} \\
3 & Historical annotation records & 80\% mapping failure; 7\% candidate note-only\newline 6\% structured-side undercapture; 7\% extraction/reference noise & \shortstack[l]{Historical sample contains\\several mismatch categories} \\
4 & Held-out mapped-row coverage & 0.6823$\rightarrow$0.7498 & \shortstack[l]{Development-derived aliases\\improve held-out coverage} \\
5 & Held-out no-overlap reduction & 39.10\%$\rightarrow$32.69\% (2{,}096$\rightarrow$1{,}752) & \shortstack[l]{Supplement reduces\\held-out no-overlap} \\
\bottomrule
\end{tabular}
\end{table*}

\subsection{Residual deterministic-pipeline error interpretation}
Among 1{,}750 full-cohort disagreements after the deterministic pipeline, the existing rule-based taxonomy assigned 1{,}694 (96.80\%) to its residual ``missing alias'' category, 30 (1.71\%) to combination or formulation mismatch, 22 (1.26\%) to lab/substance or non-medication, and 4 (0.23\%) to ambiguous abbreviation. ``Missing alias'' is the default rule category after the other checks, rather than an independently verified cause. The ten most frequent raw terms, led by temodar, avastin, carboplatin, abraxane, and doxorubicin, accounted for 577/1{,}750 disagreements (32.97\%). These frequencies identify recurring expressions for subsequent assessment of normalization and reference associations.

Overall, the analyses identify terminology, timing, mapping coverage, and possible documentation differences as contributors to cross-source mismatch; their relative prevalence remains uncertain.

\section{Discussion}
Exact same-visit overlap was low, but this did not imply widespread absence of related medication information in structured history. Many note-derived medication rows could be linked to structured records through ingredient- or category-level matching or through nearby documentation dates. The strict ontology-backed analysis further showed that residual mismatch depended in part on whether note-derived labels could be mapped to public drug concepts. These findings support reporting matching specificity and temporal proximity when comparing the two documentation sources.

On held-out patients, the measurable gain came almost entirely from lexical cleanup and curated alias coverage, while generic lexical retrieval and ontology-only matching recovered less. The contrast with generic lexical baselines points to the contribution of clinic-note-specific alias coverage. On the common MedXN-output subset, the local pipeline had higher precision and lower recall than untuned MedXN; the comparison used project-specific labels.

The strict no-overlap rate was sensitive to note-side ontology coverage. A supplement derived from development-patient mapping failures reduced strict no-overlap on held-out patients from 39.10\% to 32.69\%. The historical 100-row annotation sample records several possible explanations: mapping gaps, structured-history undercapture, extraction or reference noise, and candidate note-only evidence. These records illustrate possible causes, but do not estimate their prevalence in the corrected cohort.

Several limitations remain. The cohort comes from a private single-center clinic-note environment, and both the canonical label space and broad medication categories are project-specific. The reference relies on LLM-proposed mentions and recorded targeted corrections, with no established independent human validation. In addition, the reference layer is packet-note-based rather than an independent span-annotated corpus over every manifest note. Therefore, the main normalization endpoint should be interpreted as controlled note-side canonicalization performance within the constructed reference universe rather than as a fully independent corpus-level extraction benchmark.

The audit and alias-refinement design also impose constraints. Qwen proposed note-level medication mention strings, whereas final project-specific canonical labels were assigned downstream through deterministic projection and audit. The targeted audit is error-enriched and should not be read as global bootstrap quality. Although the alias-supplement sensitivity was rebuilt on development patients and evaluated on held-out test patients, the base deterministic alias resource remains project-curated and institution-shaped. Only Qwen2.5-32B-Instruct was used for bootstrapping; sensitivity to the choice of LLM was not evaluated. Future work should evaluate portability across institutions, extend public ontology alignment for heterogeneous note labels, and test whether the same framework transfers to other treatment-context entities beyond medications.

\section{Conclusion}
We presented a note-grounded framework for treatment-context medication normalization from clinic notes. On held-out patients, deterministic alias-aware normalization increased conditional canonical agreement from 0.8125 to 0.9358 relative to surface matching within the project-specific reference. Cross-source overlap varied with semantic and temporal matching criteria, while strict nonoverlap was sensitive to note-side ontology coverage. Clinical interpretation of these discrepancies should account for normalization coverage, matching criteria, and documentation timing.

\section*{Acknowledgment and AI Use Disclosure}
This study was partially supported by the National Cancer Institute (NCI) under grant R37CA237452.

Qwen2.5-32B-Instruct proposed note-level medication mentions for reference construction; canonical labels were assigned through downstream deterministic projection and recorded annotation corrections. An LLM assisted with polishing the manuscript wording and presentation. The authors take responsibility for the analyses, interpretations, and final text.


\begin{thebibliography}{00}

\bibitem{xu2010medex}
H. Xu, S. P. Stenner, S. Doan, K. B. Johnson, L. R. Waitman, and J. C. Denny,
``MedEx: a medication information extraction system for clinical narratives,''
\emph{J. Am. Med. Inform. Assoc.}, vol. 17, no. 1, pp. 19--24, 2010, doi: 10.1197/jamia.M3378.

\bibitem{jiang2014medexuima}
M. Jiang, Y. Wu, A. Shah, P. Priyanka, J. C. Denny, and H. Xu,
``Extracting and standardizing medication information in clinical text: the MedEx-UIMA system,''
\emph{AMIA Joint Summits Transl. Sci. Proc.}, vol. 2014, pp. 37--42, 2014.

\bibitem{uzuner2010extracting}
O. Uzuner, I. Solti, and E. Cadag,
``Extracting medication information from clinical text,''
\emph{J. Am. Med. Inform. Assoc.}, vol. 17, no. 5, pp. 514--518, 2010, doi: 10.1136/jamia.2010.003947.

\bibitem{sohn2014medxn}
S. Sohn, C. Clark, S. R. Halgrim, S. P. Murphy, C. G. Chute, and H. Liu,
``MedXN: an open source medication extraction and normalization tool for clinical text,''
\emph{J. Am. Med. Inform. Assoc.}, vol. 21, no. 5, pp. 858--865, 2014, doi: 10.1136/amiajnl-2013-002190.

\bibitem{weeks2020medextractr}
H. L. Weeks, C. Beck, E. McNeer, M. L. Williams, C. A. Bejan, J. C. Denny, and L. Choi,
``medExtractR: a targeted, customizable approach to medication extraction from electronic health records,''
\emph{J. Am. Med. Inform. Assoc.}, vol. 27, no. 3, pp. 407--418, 2020, doi: 10.1093/jamia/ocz207.

\bibitem{almeida2021twostage}
J. R. Almeida, J. F. Silva, S. Matos, and J. L. Oliveira,
``A two-stage workflow to extract and harmonize drug mentions from clinical notes into observational databases,''
\emph{J. Biomed. Inform.}, vol. 120, Art. no. 103849, 2021, doi: 10.1016/j.jbi.2021.103849.

\bibitem{turchin2009comparison}
A. Turchin, M. Shubina, E. Breydo, M. L. Pendergrass, and J. S. Einbinder,
``Comparison of information content of structured and narrative text data sources on the example of medication intensification,''
\emph{J. Am. Med. Inform. Assoc.}, vol. 16, no. 3, pp. 362--370, 2009, doi: 10.1197/jamia.M2777.

\bibitem{wang2015prescription}
Y. Wang, S. R. Steinhubl, C. deFilippi, K. Ng, S. Ebadollahi, W. F. Stewart, and R. J. Byrd,
``Prescription extraction from clinical notes: towards automating EMR medication reconciliation,''
\emph{AMIA Joint Summits Transl. Sci. Proc.}, vol. 2015, pp. 188--193, 2015.

\bibitem{nelson2011rxnorm}
S. J. Nelson, K. Zeng, J. Kilbourne, T. Powell, and R. Moore,
``Normalized names for clinical drugs: RxNorm at 6 years,''
\emph{J. Am. Med. Inform. Assoc.}, vol. 18, no. 4, pp. 441--448, 2011, doi: 10.1136/amiajnl-2011-000116.

\bibitem{newmangriffis2021ambiguity}
D. Newman-Griffis, G. Divita, B. Desmet, A. Zirikly, C. P. Ros\'{e}, and E. Fosler-Lussier,
``Ambiguity in medical concept normalization: an analysis of types and coverage in electronic health record datasets,''
\emph{J. Am. Med. Inform. Assoc.}, vol. 28, no. 3, pp. 516--532, 2021, doi: 10.1093/jamia/ocaa269.

\bibitem{ntinopoulos2025llmextract}
V. Ntinopoulos, H. Rodriguez Cetina Biefer, I. Tudorache, N. Papadopoulos, D. Odavic, P. Risteski, A. Haeussler, and O. Dzemali,
``Large language models for data extraction from unstructured and semi-structured electronic health records: a multiple model performance evaluation,''
\emph{BMJ Health Care Inform.}, vol. 32, no. 1, Art. no. e101139, 2025, doi: 10.1136/bmjhci-2024-101139.

\bibitem{song2025using}
Q. Song, J. Yang, N. Obi, C. Ni, J. L. Warner, Q. Chen, L. Song,
S. T. Rosenbloom, B. A. Malin, and Z. Yin,
``Using large language model for efficient extraction of treatment
discontinuation information,''
in \emph{Proc. Int. Conf. Artif. Intell. Med.}, 2025, pp. 379--383, doi: 10.1007/978-3-031-95841-0\_70.

\bibitem{ni2026auditable}
C. Ni, W. Liu, Q. Song, M. Murrow, B. A. Malin, and Z. Yin,
``From Clinical Free Text to Auditable Concepts: An Agentic Framework for Interpretable Prediction,''
\emph{medRxiv}, 2026, doi: 10.64898/2026.09.02.26362086.

\bibitem{lin2026graphrag}
X. Lin, Y. Feng, W. Liu, H. Ren, J. Zhou, C. Sun, and Y. Chen,
``Search Broadly, Seek Evidence on Both Sides, Decide Narrowly: Evidence-Admissible GraphRAG for Longitudinal Clinical Event Verification,''
\emph{arXiv preprint arXiv:2608.22062}, 2026.

\bibitem{zhang2024pitfalls}
J. Zhang, K. Sun, A. Jagadeesh, P. Falakaflaki, E. Kayayan, G. Tao, M. H. Ghahfarokhi, D. Gupta, A. Gupta, V. Gupta, and Y. Guo,
``The potential and pitfalls of using a large language model such as ChatGPT, GPT-4, or LLaMA as a clinical assistant,''
\emph{J. Am. Med. Inform. Assoc.}, vol. 31, no. 9, pp. 1884--1891, 2024, doi: 10.1093/jamia/ocae184.

\bibitem{wang2024applications}
L. Wang, Z. Wan, C. Ni, Q. Song, Y. Li, E. Clayton, B. Malin, and Z. Yin,
``Applications and concerns of ChatGPT and other conversational large
language models in health care: Systematic review,''
\emph{J. Med. Internet Res.}, vol. 26, Art. no. e22769, 2024, doi: 10.2196/22769.

\bibitem{eyre2021medspacy}
H. Eyre, A. B. Chapman, K. S. Peterson, J. Shi, P. R. Alba, M. M. Jones, T. L. Box, S. L. DuVall, and O. V. Patterson,
``Launching into clinical space with medspaCy: a new clinical text processing toolkit in Python,''
\emph{AMIA Annu. Symp. Proc.}, vol. 2021, pp. 438--447, 2021.

\bibitem{qwen2025technicalreport}
Qwen Team,
``Qwen2.5 Technical Report,''
\emph{arXiv preprint arXiv:2412.15115}, 2025.

\bibitem{kwon2023pagedattention}
W. Kwon, Z. Li, S. Zhuang, Y. Sheng, L. Zheng, C. H. Yu, J. E. Gonzalez, H. Zhang, and I. Stoica,
``Efficient Memory Management for Large Language Model Serving with PagedAttention,''
\emph{Proc. ACM SIGOPS 29th Symp. Operating Systems Principles}, pp. 611--626, 2023, doi: 10.1145/3600006.3613165.

\bibitem{who2026atcddd}
World Health Organization,
``The ATC/DDD Methodology,''
\emph{WHO ATC/DDD Toolkit}. [Online]. Available: https://www.who.int/tools/atc-ddd-toolkit/methodology. [Accessed: Jun. 30, 2026].

\bibitem{pathak2011ndfrt}
J. Pathak and C. G. Chute,
``Further revamping VA's NDF-RT drug terminology for clinical research,''
\emph{J. Am. Med. Inform. Assoc.}, vol. 18, no. 3, pp. 347--348, 2011, doi: 10.1136/amiajnl-2011-000161.

\bibitem{reich2024ohdsi}
C. Reich, A. Ostropolets, P. Ryan, P. Rijnbeek, M. Schuemie, A. Davydov, D. Dymshyts, and G. Hripcsak,
``OHDSI Standardized Vocabularies---a large-scale centralized reference ontology for international data harmonization,''
\emph{J. Am. Med. Inform. Assoc.}, vol. 31, no. 3, pp. 583--590, 2024, doi: 10.1093/jamia/ocad247.

\end{thebibliography}
\end{document}